# A Deep Neural Network Based Approach to Building Budget-Constrained Models for Big Data Analysis*

Rui Ming[1], Haiping Xu[2], Shannon E. Gibbs[3], Donghui Yan[4], and Ming Shao[5]
[1,2,3,5]Department of Computer and Information Science
[4]Department of Mathematics
University of Massachusetts Dartmouth, Dartmouth, MA 02747, USA
Email: {rming, hxu, sgibbbs, dyan, mshao}@umassd.edu

*Abstract*—Deep learning approaches require collection of data on many different input features or variables for accurate model training and prediction. Since data collection on input features could be costly, it is crucial to reduce the cost by selecting a subset of features and developing a budget-constrained model (BCM). In this paper, we introduce an approach to eliminating less important features for big data analysis using Deep Neural Networks (DNNs). Once a DNN model has been developed, we identify the weak links and weak neurons, and remove some input features to bring the model cost within a given budget. The experimental results show our approach is feasible and supports user selection of a suitable BCM within a given budget.

## I. INTRODUCTION

With the emergence of big data, large scale data-driven machine learning becomes increasingly important. Deep learning, also called deep structured learning, is a subfield of machine learning based on artificial neural networks (ANNs). A deep neural network (DNN) is an ANN with multiple hidden layers between the input and output layers. There are many different types of DNNs, e.g., feedforward deep neural network (FF-DNN), recurrent neural network (RNN) and convolutional neural network (CNN), all of which follow similar procedures for training and testing [1]. Deep learning approach has been very successful in recent years for processing big data from sources such as social media, Internet search engines, e-commerce platforms, and healthcare systems. Successful deep learning mechanisms require collecting a large amount of data or purchasing data from a third-party vendor on many different input features or variables in order to develop feasible and accurate models for classification and prediction. However, data collection on input features could be very expensive and time consuming. Such cost may also include preprocessing, maintenance and storage of the data associated with the input features. For example, a recommendation system of a major e-commerce application using deep learning would require storing millions of user access information per month. Dozens of features such as, the amount of time a user views a certain item, and other items that are also viewed, would be recorded for each user access. The preprocessing of such data and the costs associated with the storage, transmission and maintenance can be remarkably high. Similarly, in a deep learning application that determines when a cruise ship needs to be maintained, a huge amount of data on the status measurements and usage statistics of the different system components of the cruise ship would also be required. As one more example, in a healthcare application using deep learning, various medical test data such as blood pressure, cholesterol levels and heart rates, need to be collected to develop an accurate medical diagnose model for training and determination of certain diseases.

The cost associated with input features include the cost to collect the training and testing data as well as collection of a new data point for the classification or prediction purpose. In this study, we assume there are existing training and testing datasets for building a deep learning model. Therefore, we can focus on the total cost of collecting a new data point on all required features of the model. We call the cost for collecting a new data point the *model cost*. Note that a high model cost would also imply a high cost of acquiring the needed datasets for model training and testing. Practically, there are always limits to the budgets in deep learning applications. Due to the budget constraints, we must limit the number of features used in a model, while keeping the model accuracy high enough. In our approach, we reduce the model cost by selecting a subset of the most important features and deriving a reasonable model within a certain budget. In other words, with a given budget, we need to eliminate the least important features to ensure the model cost is lower than the budget. Since removing features typically reduces the accuracy of the model, it is required that our approach must deliver a budget-constrained model (BCM) with a reasonable accuracy. In previous work, we proposed several ways to select a set of features under a certain cost profile [2]. In this paper, we focus on deep learning methods and introduce a DNN-based approach to identifying the least important features from a DNN, subject to a given budget. Instead of deriving a single BCM, we produce a list of BCMs with expected predictive accuracies, sorted by predefined budget levels. This could be used to choose a BCM with the best predictive accuracy under a given budget, or allow a user to better trade off between budget and model accuracy.

* This material is based on work supported by 2020 University Industry Collaborative Seed Fund, University of Massachusetts.

## II. Related Work

There have been many research efforts on big data analytics using deep learning approaches. Deep architectures such as DNNs can often capture hierarchical and complex patterns of the inputs for more effective analysis of big data than traditional statistical learning methods. For example, the "Google Brain" project has used large DNNs with about one million simulated neurons and one billion simulated connections to leverage big data for image enhancement, language translation, and robotics research [3]. Esteva et al. presented deep learning techniques using DNNs for medical imaging, electronic health record data processing, and robotic-assisted surgery in the healthcare domain [4]. They also demonstrated the application of deep learning in bioinformatics, e.g., building a deep-learning system in genomics to convert raw data into input data tensors, processed by DNNs for specific biomedical applications. Xu and Gade designed a systematic approach to designing a layered knowledge graph that can be converted into a structured DNN [5]. The structured DNN model has been used for smart real estate assessments, which outperforms conventional multi-variate linear regression methods as well as prediction mechanisms used by the leading real estate companies such as Zillow and Redfin. Most of the deep learning approaches assume the availability of required datasets for predictive analysis without considering the cost associated with data collection. In contrast, our approach aims to derive budget-constrained models by eliminating the least important features.

Previous work related to cost-sensitive learning is summarized as follows. Elkan showed the proportion of negative examples in a training set would affect the optimal cost-sensitive classification decisions for problems with differing misclassification costs [6]. He recommended first developing a classifier and then using the probability estimates calculated from the classifier to compute optimal decisions. Sheng and Ling proposed a method to select a proper threshold that produces the lowest misclassification cost [7]. The experimental results showed that thresholding, as a general method to develop a cost-sensitive algorithm, has the least sensitivity on the misclassification cost ratio. O'Brien et al. analyzed the relationship between systematic errors in the class probability estimates and cost matrices for multiclass classification [8]. They explored the effect on the class partitioning of the cost matrix and demonstrated the effectiveness of learning a new partition matrix. Zhou et al. proposed a method to select features by their probabilities that are inversely proportional to the costs [9]. They constructed a decision tree with feature costs and used a random forest-based feature selection algorithm to produce low-cost feature subsets. Ji and Carin presented a formal definition of the cost-sensitive classification problem and provided a solution using a partially observable Markov decision process (POMDP) [10]. Different from traditional approaches, features were selected in a sequential manner until no additional feature acquisition could be justified based on classification results. More recently, Maliah and Shani formulated the cost sensitive classification problem as a POMDP, taking both test and misclassification costs into consideration [11]. They used a tree-based MDP approach to modeling a belief space and provided a scalable method for reasoning about future actions. Frumosu et al. proposed a method to reduce the production cost by predicting the number of faulty products while ensuring production quality delivery [12]. They reduced the problem to an imbalanced binary classification problem and solved the problem using Voronoi diagrams and the genetic algorithm.

The above cost-sensitive learning approaches provided useful methods to reduce test and misclassification costs; however, they are not aimed to provide users model options to meet the budget constraints. In addition, most of the existing cost-sensitive learning approaches are not deep learning approaches, which intrinsically have limitations in dealing with large datasets and complex problems such as medical diagnosis. In previous work [2], Yan et al. approached the problem of budget constrained learning, in terms of variable costs. They explored the solution space to produce a model schedule as a list of models, sorted by model costs and expected predictive accuracy. Based on this work, we further proposed a deep learning based approach to building budget-constrained models using deep neural networks. In this sense, our approach complements existing cost-sensitive learning approaches that are suitable for applications not involving large amount of data and provides a scalable solution to complex problems, such as cybersecurity, fraud detection and medical diagnosis.

## III. Model Cost and Budget-Constrained Models

Deep learning has been widely used in various fields such as medical diagnosis, autonomous driving, and mathematics education. DNNs are a type of deep learning methods widely adopted in big data analytics and large-scale data driven applications. Since DNN-based approaches have shown ground-breaking results in speech recognition and image recognition tasks in recent years, the number of applications using DNNs has exploded. In this paper, we demonstrate our deep learning approach using FF-DNN – a simple type of DNNs, to build BCMs for big data analysis.

### A. Model Cost of a Deep Neural Network

The FF-DNN model is usually treated as a "black box"; however, it is undeniable that every neuron in a hidden layer of a FF-DNN has certain significance or hidden semantics, and different neurons have different effects on the outputs of the model [5]. To a certain extent, the absolute weight value of a link in a neural network represents the impact of the source neuron to the target neuron. Such impact may pass through the layers of the neural network and influence the results of the output neurons. When an input neuron has the least impact on the results of the output neurons, its corresponding feature may become a candidate to be removed from the model with minimal impact on the model accuracy.

To adopt a well-trained deep learning model for prediction or classification, we need to collect data on a set of input features. For example, the set of input features to determine if a patient has a certain heart disease may include measures such as "blood pressure", "heart rate", "fasting blood sugar", "age", and "gender". The collection, purchase, and storage of data on different features may incur different feature cost. Let $F$ be the set of all measurable features in a certain domain, where $|F| = m$. Let $f = \{f_1, f_2, …, f_K\}$ be a set of input features of a model $\Phi(f)$, which uses a total of $K$ measurable features; thus, $f \subseteq F$ and $K \leq m$. Let function $c: F \mapsto Z^*$ be a mapping from feature $f \in F$ to the cost of measuring feature $f$. To simplify matters, we assume a feature cost is a nonnegative integer from the set of nonnegative integers $Z^*$. We define the model cost of $\Phi(f)$ as the summation of all feature costs as in Eq. (1).

$$C(\Phi(f)) = \sum_{i=1}^{K} c(f_i) \quad \text{where } f_i \in f \text{ and } |f| = K \quad (1)$$

Given a budget level $b$, we need to find a set of features $f \subseteq F$, such that the model cost of $\Phi(f)$ is no more than $b$, and $\Phi(f)$ has the best predictive accuracy. That is, to solve the optimal problem defined in Eq. (2).

$$\arg\max_{f \subseteq F} accuracy(\Phi(f)) \text{ subject to } C(\Phi(f)) \leq b \quad (2)$$

In our DNN-based approach, we start with all measurable features and a list of predefined budget levels. We gradually remove the least important input features until the model costs are within the budgets. For each budget level, once the least important features are removed, the remaining features form a new set of inputs for development of a new classifier. It is expected that the new model will be less accurate as the number of input features decreases; however, the costs of collecting data for training and prediction can be significantly reduced.

*B. Budget-Constrained Models*

In the context of FF-DNN, a budget-constrained model or BCM $\Phi(f)$ is defined as a 4-tuple $(S, f, w, p)$, where $S$ is the structure of the FF-DNN, $f$ is a set of input features that correspond to the set of input neurons in $\Phi(f)$, $w$ is the weights of the links in $\Phi(f)$, and $p$ the expected accuracy of the model. Note in this paper, $S$ is defined as a fully connected DNN (FCDNN), while using partially connected DNNs (PCDNNs) is envisioned as a future, and more ambitious research direction.

Given a list of budget levels $B = (b_1, b_2, …, b_n)$ and a set of measurable features $F = \{f_1, f_2, …, f_m\}$, our task is to build a list BCMs $\Phi_i$, where $1 \leq i \leq n$, and for each $\Phi_i$, with an identified subset of features from $F$ such that the model satisfies Eq. (2) for budget $b_i$. Table I shows an example of a list of BCMs with their expected model accuracy and lists of features, sorted by budget levels. With a given budget and a required accuracy, we can find the most suitable model from the table. For example, if the given budget is 1750, and the required model accuracy is 0.94, we shall choose the BCM with the set of features {1,4,5,8,10}, whose model cost is 1500 that is less than 1750.

TABLE I. AN EXAMPLE OF A LIST OF BCMs UNDER BUDGETS

| Model | Budget | Accuracy | Features |
|---|---|---|---|
| $\Phi_1$ | 3000 | 0.9615 | {1,2,4,5,8,10,11,12} |
| $\Phi_2$ | 2500 | 0.9519 | {1,2,4,5,8,10,11} |
| $\Phi_3$ | 2000 | 0.9433 | {1,4,5,8,10,11} |
| $\Phi_4$ | 1500 | 0.9406 | {1,4,5,8,10} |
| $\Phi_5$ | 1000 | 0.9357 | {1,5,8,10} |
| $\Phi_6$ | 500 | 0.9325 | {1,5,10} |

*C. A Framework for Building a Budget-Constrained Model*

With sufficient training and testing datasets, our approach aims to develop a BCM with its model cost within a given budget level. The framework for generating a BCM under a given budget is illustrated in Fig. 1.

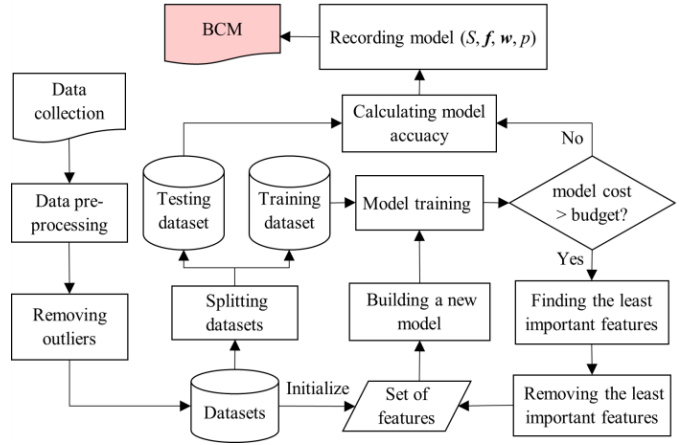

Fig. 1. A Framework of generating a BCM under a given budget

For any raw data, whether captured by measurement or purchased from a third-party vendor, there is typically a lot of unnecessary information. We first need to preprocess the data and retrieve the needed fields in a desired format. Since data points with missing information or wrong information could negatively affect the training and testing results, such data points must be fixed or considered as outliers to be removed from the dataset. The dataset is then split into a training dataset and a testing dataset. Note that to simplify Fig. 1, we do not show the process of partitioning the dataset into $k$ equal sized subsamples for the $k$-fold cross-validation purpose. We extract all the features from the dataset to build the first model. After the model is fully trained, we check if the model cost is higher than the given budget. If the answer is yes, we find the least important feature, remove it using an algorithm described in Section IV, and create a new model using the remaining features. This procedure is repeated until the model cost becomes less than or equal to the given budget. In this case, the testing dataset is used to calculate the expected model accuracy. Finally, the 4-tuple $(S, f, w, p)$, i.e., the structure $S$ of the FF-DNN, the set of input features $f$, the weights of the links $w$ in the model, and the expected accuracy $p$ of the model, is recorded as the resulting BCM for the given budget.

IV. GENERATION OF BUDGET-CONSTRAINED MODELS

*A. Identifying the Least Important Input Feature*

In our DNN-based approach, we define a set of thresholds for the links of the neural network, designed to identify and eliminate the weak links. When a neuron's output links are all identified as weak links, the neuron is considered to have minimal impact on the output, and thus, it is considered a weak neuron. Our approach starts with the last hidden layer that most directly affects the output neurons, and then works backward to determine the weak links and weak neurons. The procedure repeats until we find weak input neurons, whose corresponding features become a candidate to be removed from the model.

We now use a few examples to show how to identify weak links and weak neurons. Since multiple source neurons link to a target neuron, the weights of each link represent their impact on the target neuron. The higher absolute value of a weight, the higher the impact a source neuron has on the target neuron. We can identify the weak links by setting a threshold for each target neuron of a link. For example, in Fig. 2 (a), the link threshold for target neuron $t_n$ is set to 0.3. Consequently, the link from source neuron $n_1$ to $n$ is marked as a weak link (denoted by a dashed line) as the weight of the link is 0.1 that is less than $t_n$.

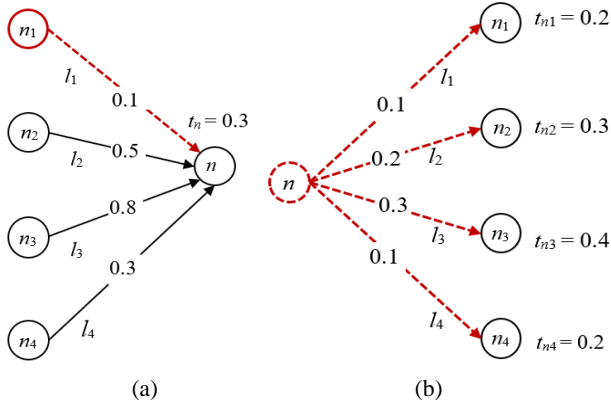

(a)                                              (b)

Fig. 2. Examples of (a) weak link and (b) weak neuron

On the other hand, a source neuron links to multiple target neurons. If the links coming from a source neuron have all been identified as weak links, that source is marked as a weak neuron. For example, in Fig. 2 (b), all links coming from source neuron $n$ are weak links because their link weights are less than their corresponding thresholds; thus, neuron $n$ is marked as a weak neuron denoted by a dashed circle.

If a neuron is identified as a weak neuron, its impact on the outputs of the neural network is considered minimal. Therefore, all links connecting that weak neuron are considered as weak links because if we remove the weak neuron from the DNN, all its incoming links will also be removed. Fig. 3 shows such an example with neuron $n_5$ being a weak neuron.

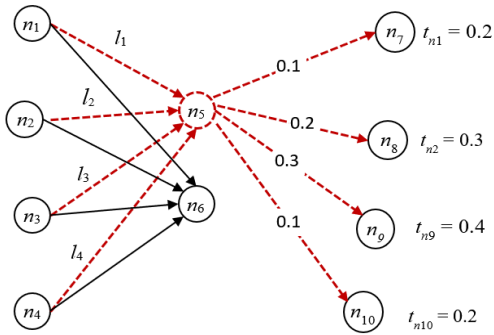

Fig. 3. An example of weak links connecting to a weak neuron

As shown in Fig. 3, since neuron $n_5$ is a weak neuron, neuron $n_1$ to $n_4$ would have little impacts on the outputs of the neural network through their links to $n_5$; therefore, we can reasonably mark links $l_1$, $l_2$, $l_3$ and $l_4$ as weak links. Being said, a link is marked as a weak link in either of the following two cases: 1) its weight is less than the threshold, and 2) its target neuron is a weak neuron.

Fig. 4 presents an example of a FF-DNN model with four layers including two hidden layers. There are three input neurons $n_{11}$, $n_{12}$ and $n_{13}$, which correspond to three input features. All neurons except the input neurons have been assigned thresholds. Note that the thresholds for the neurons can be different, and each threshold of a neuron $n$ is initialized based on the weights of all links that connect to neuron $n$. As described later in this section, the thresholds need to be adjusted if no weak input neuron can be identified.

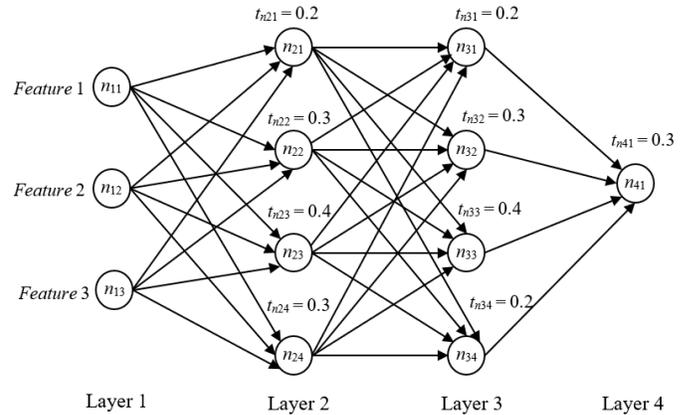

Fig. 4. An example of a FF-DNN model

The steps to identify weak links and weak neurons of the neural network in Fig. 4 are illustrated in Fig. 5. From the figure, we can see the process starts with the last hidden layer and works backward to the input layer. For example, in Layer 3, since the neuron $n_{31}$ contains only one output link, which is a weak link, it is marked as a weak neuron. Similarly, in Layer 2, since neuron $n_{21}$ contains links that are either weak or connect to a weak neuron, neuron $n_{21}$ is then marked as a weak neuron as well. Finally, in Layer 1, two input neurons $n_{11}$ and $n_{12}$ are identified as weak neurons; thus, their corresponding input features are candidate features to be removed from the model. It is worth noting that, in our approach, when more than one weak input neurons are identified, the least important feature is considered to be the one having the highest feature cost; therefore, minimizing the model cost.

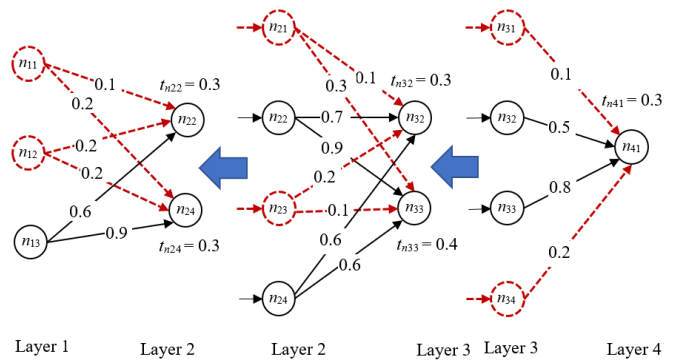

Fig. 5. The steps to identify weak links and weak neurons

The procedure of finding the least important feature is shown in Algorithm 1. As described in the algorithm, given a FF-DNN $\Phi(f)$ with $L$ layers, all neurons and links in $\Phi(f)$ are initially considered strong. A very low initial threshold $\delta_n$ is set for each neuron $n$, except for the input layer, based on the weights of the input links to neuron $n$. Starting from the last hidden layer $l_{L-1}$, all the weak neurons and weak links are marked in a backward manner. To ensure that the input layer contains at least one weak neuron, the value of each threshold can be increased gradually. Finally, a weak neuron in the input layer is selected and its corresponding input feature is identified as the least important feature $f^*$.

---

**Algorithm 1: Finding the Least Important Feature**

**Input:** A FF-DNN $\Phi(f)$ with $L$ layers including input layer $l_1$ and output layer $l_L$, where $f$ is a set of features and $L \geq 4$.
**Output:** The least important feature $f^*$.

1. Let all neurons/links in $\Phi(f)$ be strong neurons/links
2. Let $f^*$ be the least important feature, initialized to *null*
3. **for** $i = 2$ to $L$
4.   **for** each neuron $n$ in layer $l_i$
5.     Initialize the threshold $t_n$ of neuron $n$ with small value $\delta_n$
6. **while** $f^*$ is *null*
7.   **for** $i = L-1$ to 1 // identify weak links/neurons backward
8.     **for** each target neuron $\beta$ in $l_{i+1}$
9.       **for** each source neuron $\alpha$ in $l_i$
10.         Let $w_\gamma$ be the weight of the link $\gamma$ from $\alpha$ to $\beta$
11.         **if** $w_\gamma < t_\beta$ **or** $\beta$ is a weak neuron
12.           Mark link $\gamma$ as a weak link
13.       **for** each source neuron $\alpha$ in $l_i$
14.         **if** all links from source neuron $\alpha$ are weak links
15.           Mark source neuron $\alpha$ as a weak neuron
16.   **if** there is no weak neuron in input layer $l_1$
17.     Let threshold $t_n$ of each target neuron $n$ in $\Phi(f)$ be $2*\delta_n$
18.   **else** // there are one or more weak input neurons
19.     Select a weak neuron $\alpha^*$ in $l_1$ with highest feature cost
20.     Set $f^*$ to the input feature corresponding to $\alpha^*$
21. **return** $f^*$

---

*B. Generating a FF-DNN based BCM*

Once we are able to identify the least important input feature in our FF-DNN based deep learning approach, we can generate a FF-DNN model that satisfies a budget requirement. Let a given budget be $b$. We develop a FF-DNN model that satisfies the requirements described in Eq. (2). This may require going through a number of steps to remove more than one input feature to meet the budget requirement. Each time when the least important feature is removed, we build a new FF-DNN model and train it using the same datasets. It is expected that the new model is less accurate than its previous model version as the number of input features decreases. With the trained new model, we identify the least important feature again until the budget requirement is met. Algorithm 2 shows the procedure to generate a BCM given budget level $b$, dataset $D$ with a set of features $F$, and model cost function $C(\Phi(f))$. To make the model cost $C(\Phi(f)) \leq b$, starting from $f = F$, the method gradually removes the least important feature using Algorithm 1. Finally, the model $\Phi(f)$ is created and trained on the $f$ that satisfies $C(\Phi(f)) \leq b$, and the corresponding 4-tuple $(S, f, w, p)$ representing BCM $\Phi(f)$ is returned as the result.

---

**Algorithm 2: Generating a BCM Under a Given Budget**

**Input:** Dataset $D$ with a set of $m$ measurable features $F = \{f_1, f_2, \ldots, f_m\}$, model cost function $C(\Phi(f))$, and a given budget $b$.
**Output:** 4-tuple $(S, f, w, p)$ representing BCM $\Phi(f)$ with $C(\Phi(f)) \leq b$.

1. Let $f$ be the set of measurable features $F$
2. Randomly partition $D$ into $k$ equal sized subsamples.
3. **while** $C(\Phi(f)) > b$ // model cost is greater than the given budget
4.   Create a FF-DNN $\Phi(f)$ with a set of features $f$
5.   Train and test $\Phi(f)$ with dataset $D$ using $k$-fold cross-validation
6.   Invoke Algorithm 1, and let $f^*$ be the least important feature
7.   Remove feature $f^*$ from $f$
8. Create a FF-DNN $\Phi(f)$ with a set of features $f$
9. Train and test $\Phi(f)$ with $D$, and save weights $w$ and accuracy $p$
10. Let $S$ be the structure of FF-DNN $\Phi(f)$.
11. **return** 4-tuple $(S, f, w, p)$

---

*C. Generating a List of BCMs*

Developing a deep learning model under a specific budget may possibly result in failing to achieve the required predictive accuracy or wasting money on unnecessary features. For example, a low given budget for a deep learning model adopted in a cardiac diagnosis application may only use a limited number of features, which could make the prediction accuracy less than 60%. Such an application is obviously not marketable. On the other hand, suppose a vehicle routing simulation application has already achieved close to 100% prediction accuracy with a reasonable model cost. If we continue to improve the model with more features under a higher budget, it cannot improve the predictive accuracy significantly and will inevitably waste money. To avoid the above undesirable situations, users shall be allowed to trade off between various budget levels and the required predictive accuracy for a suitable cost-effective deep learning model. Algorithm 3 shows the procedure to generate a list of BCMs $L_{BCM}$, given a maximum budget $b_{max}$ and a distance $d$ between two consecutive budget levels. Each generated BCM satisfies the minimal accuracy requirement as well as its corresponding budget requirement.

---

**Algorithm 3: Generating a List of BCMs**

**Input:** The maximum budget $b_{max}$, the distance $d$ between two consecutive budget levels, and the minimum required predictive accuracy $p_{min}$
**Output:** A list of BCMs $L_{BCM}$ that meet the budget and predictive accuracy requirements

1. Let $L_{BCM}$ be a list of BCMs, initialized to an *empty* list.
2. Let $b$ be a budget level, initialized to $b_{max}$
3. **while** $b > 0$ // a budget level should be always greater than 0
4.   Invoke Algorithm 2, and let $(S, f, w, p)$ be a 4-tuple representing BCM $\Phi(f)$ with $C(\Phi(f)) \leq b$
5.   **if** $p \geq p_{min}$ // the expected model accuracy is no less than $p_{min}$
6.     Add the 4-tuple $(S, f, w, p)$ into $L_{BCM}$
7.   **else return** $L_{BCM}$
8.   $b = b - d$
9. **return** $L_{BCM}$

## V. Case Study

The major goal of our approach is to develop a list of DNN models that meet budget requirements, while keeping the predictive accuracy of each model as high as possible. To validate the feasibility and performance of our approach, we conduct experiments on two datasets from the UC Irvine Machine Learning Repository [13]. The two datasets are the *Early-stage diabetes risk prediction dataset* (*DS*1) and the *Heart disease dataset* (*DS*2). To facilitate the application of our approach and ensure fully trained models as well as improved model accuracy, we adopt the TensorFlow [14] to develop FF-DNNs for the experiments and apply *k*-fold cross-validation to train and test the models. Each dataset is randomly divided into 10 datasets for 10-fold cross-validation. For categorical data, we use one-hot encoding to divide the corresponding feature into multiple in order to improve model performance. For example, the feature *Itching* in *DS*1 is a categorical feature with a value of either "Yes" or "No", representing the presence or absence of itching symptoms, respectively. Using the one-hot encoding, the feature *Itching* can be split into two features, namely *Itching_Yes* and *Itching_No*. If *Itching* has value "Yes", it is replaced by two features *Itching_Yes* = 1 and *Itching_No* = 0; otherwise, we set *Itching_Yes* = 0 and *Itching_No* = 1.

### A. The Early Stage Diabetes Risk Prediction Dataset

The Early-stage diabetes risk prediction dataset includes 520 instances, collected using direct questionnaires from the patients of Sylhet Diabetes Hospital in Sylhet, Bangladesh [13]. There are 13 categorical attributes used as features, namely *Age*, *Sex*, *Polydipsia*, *Weakness*, *Polyphagia*, *Genital thrush (Gthrush)*, *Visual blurring (Vblur)*, *Itching*, *Irritability*, *Delayed healing (Dheal)*, *Partial paresis (Par)*, *Muscle stiffness (Mstiff)* and *Alopecia.* Each input feature is assigned a feature ID as shown in Table II.

TABLE II. FEATURES IN EARLY-STAGE DIABETES RISK PREDICTION DATASET

| *DS*1 Input Feature [Feature ID] | | | | |
|---|---|---|---|---|
| Dheal [1] | Alopecia [2] | Vblur [3] | Obesity [4] | Itching [5] |
| Gthrush [6] | Polydipsia [7] | Irritability [8] | Polyphagia [9] | Par [10] |
| Mstiff [11] | Weakness [12] | Age [13] | | |

The label of each data point is an output categorical feature of *Diabetes*, which has the value of either "Yes" or "No", indicating whether a patient has diabetes or not. The FF-DNN models that we build for this dataset have 5 hidden layers with 120 hidden neurons in each hidden layer. We set the feature costs randomly by sampling from [100, 300] uniformly, except for the costs of *Sex* and *Age*, which are set to 0. In the following experiments, the maximum budget level $b_{max}$ is set to 1900, which is greater than the total cost of all features in the diabetes dataset; thus, the initial BCM model shall consist of all the features with the potential maximum predictive accuracy. We set a distance $d = 200$ between two consecutive budget levels, gradually decrease the budget level, and derive the corresponding BCMs. This process stops when the predictive accuracy becomes less than the minimum required predictive accuracy $p_{min} = 0.65$, as predefined for the experiments.

Table III shows a list of BCMs generated by applying Algorithm 3, where the features are represented by the feature ID as defined in Table II. For example, the BCM $\Phi_8$, with the given budget of 500 and expected predictive accuracy of 0.8173, has a set of input features {3,6,7,13}, representing the features of *Visual blurring*, *Genital thrush*, *Polydipsia*, and *Age*.

TABLE III. LIST OF BCMs FOR THE EARLY-STAGE DIABETES DATASET

| Model | Budget | Accuracy | *DS*1 Input Features |
|---|---|---|---|
| $\Phi_1$ | 1900 | 0.9615 | {1,2,3,4,5,6,7,8,9,10,11,12,13} |
| $\Phi_2$ | 1700 | 0.9423 | {1,3,4,5,6,7,8,9,10,11,13} |
| $\Phi_3$ | 1500 | 0.9423 | {1,3,4,5,6,7,8,9,10,13} |
| $\Phi_4$ | 1300 | 0.9327 | {1,3,5,6,7,8,9,10,13} |
| $\Phi_5$ | 1100 | 0.9327 | {1,3,6,7,9,10,13} |
| $\Phi_6$ | 900 | 0.9135 | {3,6,7,9,10,13} |
| $\Phi_7$ | 700 | 0.8462 | {3,6,7,9,13} |
| $\Phi_8$ | 500 | 0.8173 | {3,6,7,13} |
| $\Phi_9$ | 300 | 0.7115 | {3,13} |

The list of BCMs in Table III allows a user to select an appropriate deep learning model based on the budget and accuracy requirements. For example, when the given budget is 1600 and the required accuracy is 0.94, the user shall select the BCM $\Phi_3$ with the set of features {1,3,4,5,6,7,8,9,10,13}. In this case, the expected predictive accuracy is 0.9423, which is greater than the required accuracy 0.94. However, if the required accuracy becomes 0.95, the user will have to increase the budget to 1900 and select the first model $\Phi_1$ with expected predictive accuracy 0.9615, being greater than 0.95.

To demonstrate the expected performance of our approach, we compare it with two different approaches: the cost-based approach and the random selection approach. With the given features and the cost function, the cost-based approach works in accordance with the principle that it always removes the most expensive feature to make the model cost decrease quickly; while the random selection approach randomly removes a feature each time to reduce the model cost.

For each of the three approaches, we generate 10 BCMs for each budget level, and select the model with the highest prediction accuracy. The highest predictive accuracy vs. model cost at each budget level is presented in Fig. 6.

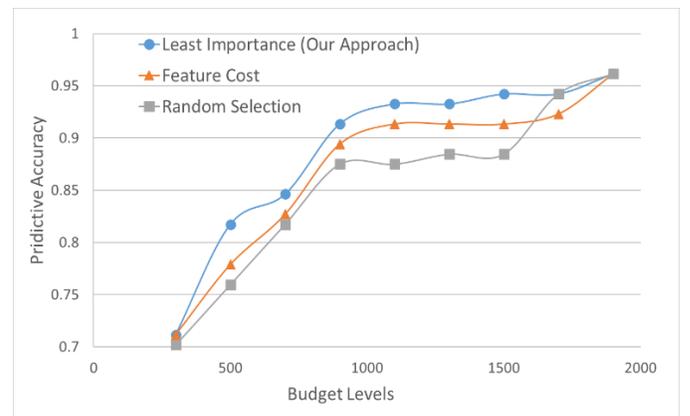

Fig. 6. Predictive accuracy over predefined budget levels (*DS*1)

From Fig. 6, we can see that our approach outperforms the other two approaches at each budget level, while the cost-based approach works better than the random-selection approach in most of the cases. Since the cost-based approach removes the most expensive feature at each step, it can remove the minimum number of features to make the model cost below a given budget. Compared with the random-selection approach, the cost-based approach would generally perform better than the random selection approach as more features could be kept for a given budget, potentially leading to a higher accuracy. However, the cost-based approach may also mistakenly remove the most expensive feature that is also an important one. This is the reason why the cost-based approach cannot perform as well as our approach. Note that our approach always removes the least important feature first, which has the lowest impact on the model prediction accuracy. In the figure, all three curves intersect at budget level 1900, the reason being, all three approaches share the same FF-DNN that uses all input features, and thus, they have the same accuracy. In addition, we notice that the accuracies of all three methods drop sharply when the budget level becomes less than 900; whereas the budget levels above 900 maintain high accuracies of all three methods. This indicates that the BCM $\Phi_6$ from Table III with the budget level of 900 may be considered as the most cost-effective model.

To further demonstrate that our approach leads to a higher degree of model accuracy than the other two approaches. We conduct experiments using the three approaches by removing only one feature at a time. Fig. 7 shows the comparison results among the three approaches showing how accuracy changes with the number of features removed. As demonstrated in the figure, for any number of features removed, our approach consistently achieves the highest model accuracy than the other two approaches.

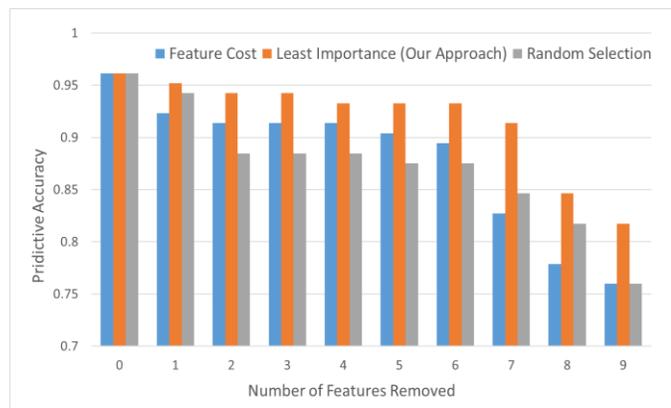

Fig. 7. Accuracy changes with the number of features removed (*DS*1)

B. The Heart Disease Dataset

The Heart disease dataset contains 76 attributes, but only 14 features is used in this experiment for demonstration purpose [13]. The 14 features include 7 categorical attributes, namely *Sex*, *Chest pain type (Cp)*, *Slope of the peak exercise ST segment (Slope)*, *Resting electrocardiographic results (Restecg)*, *Number of major vessels colored by fluoroscopy (Ca)*, *Exercise induced angina (Thal)*, *Thallium Stress Test (Exang)*, along with 6 integer attributes, namely *Age*, *Resting blood pressure (Trestbps)*, *Serum cholestoral in mg/dl (Chol)*, *Fasting blood sugar (Fbs)*, *Maximum heart rate achieved (Thalach)*, and *ST depression induced by exercise relative to rest (Oldpeak)*. Each input feature is assigned a feature ID as shown in Table IV.

TABLE IV. FEATURES IN HEART DISEASE DATASET

| *DS*2 Input Feature [Feature ID] | | | | |
|---|---|---|---|---|
| Ca [1] | Exang [2] | Cp [3] | Thal [4] | Thalach [5] |
| Oldpeak [6] | Trestbps [7] | Slope [8] | Fbs [9] | Restecg [10] |
| Chol [11] | Sex [12] | Age [13] | | |

The label of each data point is an output categorical feature of *Diagnosis of heart disease*, which has the value of either "Yes" or "No", indicating whether a patient has a heart disease or not. The FF-DNN model we built for this dataset contains 3 hidden layers with 200 hidden neurons in each hidden layer. Similar to the experiments on the Early-stage diabetes risk prediction dataset, we set the maximum budget level $b_{max}$ to 1600, the distance $d = 200$ between two consecutive budget levels, and the minimum required predictive accuracy $p_{min}$ to 0.65. Table V shows the list of BCMs generated using our approach with random feature costs sampling from [100, 300].

TABLE V. LIST OF BCMs FOR THE HEART DISEASE DATASET

| Model | Budget | Accuracy | *DS*2 Input Features |
|---|---|---|---|
| $\Phi_1$ | 1600 | 0.9333 | {1,2,3,4,5,6,7,8,9,10,11,12,13} |
| $\Phi_2$ | 1400 | 0.9111 | {1,2,3,4,6,8,9,10,11,12,13} |
| $\Phi_3$ | 1200 | 0.9111 | {1,2,3,6,8,9,10,11,12,13} |
| $\Phi_4$ | 1000 | 0.8889 | {1,2,3,6,9,10,12,13} |
| $\Phi_5$ | 800 | 0.8444 | {1,2,6,9,10,12,13} |
| $\Phi_6$ | 600 | 0.8222 | {2,6,10,12,13} |
| $\Phi_7$ | 400 | 0.8000 | {6,10,12,13} |
| $\Phi_8$ | 200 | 0.7778 | {6,12,13} |

Now, we compare the performance of our approach with that of the cost-based approach and the random selection approach by generating lists of BCMs for various budget levels. The results of predictive accuracy over predefined budget levels for the three approaches are shown in Fig. 8.

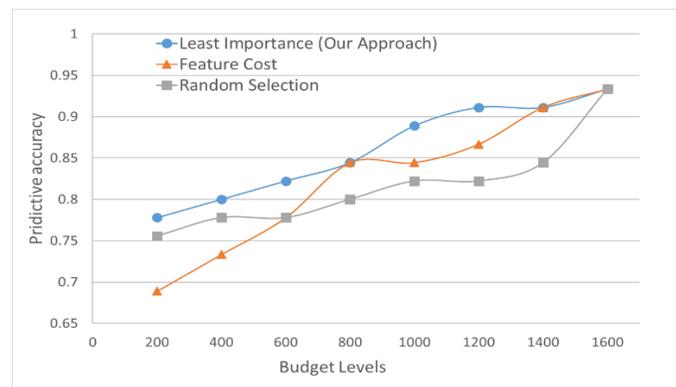

Fig. 8. Predictive accuracy over predefined budget levels (*DS*2)

From Fig. 8, we can see that our approach has the highest model accuracy at all budget levels, while the cost-based approach has higher accuracy than the random selection approach at most of the budget levels. These results are consistent with those from the previous experiments on $DS1$. However, we notice that when the two features *Exang* and *Restecg* are removed from BCM $\Phi_6$ and $\Phi_7$, respectively, the predictive accuracy is not significantly changed. This is different from the situation shown in Fig. 6, where accuracy drops sharply when the budget level becomes low. Since our approach always removes the least important feature first, the features *Exang* and *Restecg* are supposed to be important ones; thus, removing them shall result in significant decrease of the predictive accuracy. The reason why this does not happen could be explained by the correlations the input features may have with each other. In this particular case, the importance of the features *Exang* and *Restecg* may have relied on features that have been removed, e.g., features *Ca* and *Fbs* in BCM $\Phi_5$.

Similar to the experiments on $DS1$, we develop models using the three approaches by removing only one feature at a time. Fig. 9 shows the comparison results among the three approaches. As shown in the figure, for any number of features removed, our approach again consistently achieves the highest model accuracy than the other two approaches.

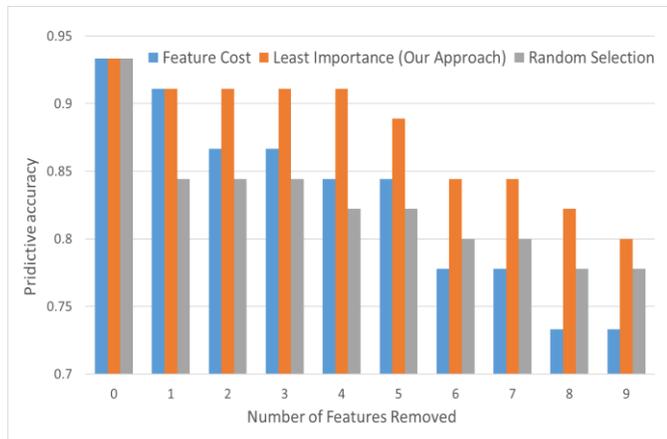

Fig. 9. Accuracy changes with the number of features removed ($DS2$)

## VI. CONCLUSIONS AND FUTURE WORK

Big data analytics is increasingly becoming one of the trending industry practices, but it has also brought major challenges for data processing, data maintenance and accurate prediction. One such major challenge is associated with the high cost of model features in many applications. In this paper, we introduced a DNN-based approach to developing deep learning models subject to budget constraints. Our approach can gradually reduce the model cost by removing the least important feature at each step. We present an algorithm to find weak links and weak neurons in a backward manner and identify the least important feature in a model. To support user selection of a suitable BCM under a given budget, or trade off between budget and predictive accuracy, we demonstrate how to generate a list of BCMs under predefined budget levels and a minimum required accuracy. Since our approach is based on deep neural network, it is scalable and provides a promising method for big data analysis.

In our current work, we performed experiments using the FF-DNN on standard datasets. In future work, we will adopt more advanced DNNs such as RNN, further verify the performance of our approach using much larger datasets, and evaluate the computational cost of our approach. We will also look into the dependency among input features, and seek a more efficient method by removing a group of highly correlated but less important features. Instead of deriving a list of BCMs, we will explore to build dynamic models with mutable feature costs. This would require developing real-time classifiers as shown in previous work [15]. Finally, we plan to build partially connected FF-DNNs under given budget levels. This could be a challenging task because partially connected FF-DNNs are currently not supported in major deep learning tools such as TensorFlow. However, as in earlier work [5], using partially connected DNNs in our deep learning approach can simplify the computation process and lead to more efficient BCMs.